\documentclass[wcp]{jmlr}

\usepackage{longtable}

\usepackage{booktabs}
\usepackage{algorithm}
\usepackage{algorithmic}
\usepackage{multirow}

\jmlrvolume{60}
\jmlryear{2016}
\jmlrworkshop{ACML 2016}

\title[RL for parallelization]{Reinforcement Learning Approach for Parallelization in Filters Aggregation Based Feature Selection Algorithms}

  \author{\Name{Ivan Smetannikov} \Email{ismetannikov@corp.ifmo.ru}\\
  \Name{Ilya Isaev} \Email{isaev@rain.ifmo.ru}\\
  \Name{Andrey Filchenkov} \Email{afilchenkov@corp.ifmo.ru}\\
   \addr ITMO University, St. Petersburg, Kronverksky Pr. 49}

\begin{document}

\maketitle

\begin{abstract}
One of the classical problems in machine learning and data mining is feature selection. A feature selection algorithm is expected to be quick, and at the same time it should show high performance. MeLiF algorithm effectively solves this problem using ensembles of ranking filters. This article describes two different ways to improve MeLiF algorithm performance with parallelization. Experiments show that proposed schemes significantly improves algorithm performance and increase feature selection quality.
\end{abstract}

\section{Introduction}
Almost all business and scientific problems nowadays involve processing huge amounts of data with machine learning algorithms. Due to its universal applicability, machine learning became one of the most promising and researched scientific domains. In particular, it has application in bioinformatics~\citep{bolon2014review,saeys2007review}, as giant amounts of data about gene expression of different organisms are obtained in this field. In order to filter data noise and reduce model complexity, it is necessary to select the most relevant features. Techniques and methods achieving this goal are called feature selection. Gene expression data can enable researchers to spot which DNA pieces are responsible for reactions to particular environment change or some internal processes of an organism. The main problem met in processing such data is the high dimensionality of instances. Gene expression datasets often have a high number of features and relatively low number of objects. For a dataset with these properties, it is very hard to build a model that fits the data well.

A feature selection algorithm meets several requirements. It is expected to work fast and show good performance. However, no universal algorithm for feature selection exists. Wrappers~\citep{kohavi1997wrappers} are the family of methods based on searching for an optimal feature subset that maximizes preselected classifier effectiveness. Such problem statement leads to high performance of a found solution. However, the size of search space grows exponentially of the instance dimensionality. This fact makes wrapper rarely applicable in bioinformatics, as the number of features in datasets could be up to hundreds of thousands. In these cases, other feature selection algorithms known as filters~\citep{sanchez2007filter} are used. Filters are based on estimation of feature importance. Filters usually perform worse than wrappers, but they are much faster. A special group of feature selection methods are embedded selectors~\citep{lal2006embedded} that uses particular properties of a selected classifier.

Ensembling, which is the process of building a combination of several simple algorithms, is a widely used technique in machine learning~\citep{bolon2012ensemble}. MeLiF algorithm proposed in~\citep{smetannikov2016melif}, applies ensembling to feature selection. This algorithm tries to find such linear combination of basic ranking filters, which selects the most relevant features of the dataset. Ranking filter itself consists of two separate parts: a feature importance measure and a cutting rule. Basically, MeLiF tries to tune coefficients of feature importance measure linear combination. This process involves classifier training, evaluating and comparing with ranking filters themselves, thus making it comparatively slow. This is why parallelization can become really handy and helpful to improve algorithm computational time.

The simplest parallelization scheme called MeLiF+ is described in~\citep{isaev2016melif+}. The main disadvantage of this na\"ive scheme is that it does not scale well. It starts search of the best coefficient vector from several starting points using a separate thread for each point. When one of these optimization processes ends, this thread just stops and its resources stay unreleased therefore they cannot be used for further work. Thus, it is not useful to allocate a lot of resources for this process as most of them will stay unused.

To overcome this problem, it is necessary to use cores of processing server more effectively. While processing, MeLiF visits a lot of points in the linear space, so we can process points using a task executor. This research proposes two different approaches to using parallel coordinate descent in building ensembles of ranking filters called PQMeLiF and MAMeLiF. The first algorithm stores points that should be processed in a priority queue. The second algorithm solves theparallelization problem by reducing it to the multi-armed bandit problem.

The remainder of the paper is organized as follows: Section 2 describes MeLiF algorithm, Section 3 contains the proposed parallelization schemes, Section 4 outlines experimental setup, Section 5 contains experiment results, and finally Section 6 contains conclusion.

This paper is a version of the paper accepted to 5th International  Young Scientists Conference in HPC and Simulation.

\section{Linear combination of ranking filters}

Ranking filter $f$ is a pair $\langle m, \kappa\rangle,$ where $m$ is a feature importance measure and $\kappa$ is a cutting rule. For each object feature, $m$ return its importance for label prediction. For a sorted list of features, $\kappa$ cuts the worst. The core idea of MeLiF is to use several ranking filters $f_1, \ldots, f_N$ in order to merge them into a single ranking filter by finding the most effective linear combination of their feature importance measures. This combination is a new feature importance measure, while the cutting rule can be inherited from all the ranking filters (usually, it is chosen empirically). Any performance measure may be used to evaluate a ranking filter effectiveness. In this paper, we use classifier effectiveness estimated with $F_{1}$ score.

Thus, MeLiF simply optimizes a function in the $N$-dimensional space, where $N$ is the number of basic ranking filers. Evaluation of this function is comparatively costly (we need to run a classifier). However, dimensionality of the search space is less by an order of magnitude than the number of features. This detail allows classifying an algorithm from MeLiF family as a hybrid of filter and wrapper, which inherits filter speed and wrapper focus on resulting performance.

The algorithm is parametrized with the following hyperparameters:
\begin{itemize}
\item $\delta \in \mathbb{R}$, a value of grid spacing;
\item $P \in 2^N$, starting points;
\item $evaluate$, the function for classifier effectiveness evaluation at given point in the search space.
\end{itemize}

The original MeLiF performs coordinate descent in the search space. It has been observed during experiments that the best option is this particular choice of starting points: $(1,0,\dots,0),$ $(0,1,\dots,0),\dots ,(0,0,\dots,1)$ corresponding to the only one basic ranking filter used, and $(1,1,\dots,1)$ corresponding to the equally weighted combination of all the basic ranking filters.

For each point it reaches, MeLiF tries to shift each coordinate value to $+\delta$ and $–\delta$. Then it evaluates effectiveness of each point. If evaluation result is greater than the current maximum, the algorithm assigns the current maximum to be equal to the coordinates of this point and starts searching from its first coordinate. If all coordinates are shifted to $+\delta$ and $-\delta$ and no quality improvement is observed, then the algorithm stops.

\begin{algorithm}
\caption{MeLiF pseudo code}
\begin{algorithmic} 
\REQUIRE starting points, $\delta$, $evaluate$
\STATE $q^{*} = 0$
\STATE $p^{*}$
\FOR {$p:points$}
    \STATE $q=evaluate(p)$
    \IF{$q>q^{*}$}
        \STATE $p^{*}=p$
        \STATE $q^{*}=q$
    \ENDIF
\ENDFOR

\STATE $smthChanged=true$
\WHILE{$smthChanged$}
    \FOR{$dim:p.size$}
        \STATE $p+=p\{p[dim]+\delta\}$
        \STATE $q+=evaluate(p+)$    
        \IF{$q+>q^{*}$}
            \STATE $q^{*} = q+$
            \STATE $p^{*} = p+$
            \STATE $smthChanged=true$
            \STATE $break$
        \ENDIF
        \STATE $p-=p\{p[dim]-\delta\}$
        \STATE $q-=evaluate(p-)$
        \IF{$q->q^{*}$}    
            \STATE $q^{*} = q+$
            \STATE $p^{*} = p+$
            \STATE $smthChanged=true$
            \STATE $break$
        \ENDIF
    \ENDFOR
\ENDWHILE
\RETURN $p^{*}, q^{*}$
\end{algorithmic}
\end{algorithm}

For each point obtained during the coordinate descent, the algorithm measures the value of the resulting linear combination of basic filters for each feature in the dataset. After that, the results are sorted, and the algorithm selects $m$ topmost features. They are used to train and test a particular classifier. The classification quality is treated as point score. It is cached and compared to other points.

\section{MeLiF parallel optimizations}

In this paper, we propose two parallelization schemes of MeLiF and show that some of their configurations have speed improvement growing linearly of the number of processors. Furthermore, the proposed schemes show equal or even better performance quality in comparison with the single-threaded version of MeLiF.

The first proposed parallelization scheme is named PQMeLiF. In this name, “PQ” stands for “priority queue”. This algorithm is a variation of the best-first search algorithm~\citep{russell2009artificial}. The algorithm stores points that should be processed in a priority queue. On each iteration, the algorithm polls a point from the queue, calculates its score, and puts all its not visited neighbors back to the queue with the priority equal to the calculated score. Before initiating the algorithm, starting points should be put into a queue with maximum priority, which is $1.0$. This ensures that all of them will be processed at the very beginning, so all the starting points will be taken into account simultaneously.

Unlike MeLiF and MeLiF+, PQMeLiF enables to tune halting criteria to find a trade-off between feature selection quality and algorithm performance: we can limit the number of points that should be processed. Experiments determined an optimal number that enables the algorithm to perform better than the original MeLiF, but as fast as possible.

\begin{algorithm}
\caption{PQMeLiF pseudo code}
\begin{algorithmic} 
\REQUIRE starting points, $\delta$, $evaluate$, $T$
\STATE $q = PriorityBlockingQueue$
\STATE $p^{*}$
\FOR {$p:points$}
    \STATE $enqueue(q,p,1.0)$
\ENDFOR

\FOR{$each\ new\ thread\ run\ in\ T\ threads$}
    \WHILE{$!mustStop()$}
        \STATE $p=dequeue(q)$
        \STATE $score=evaluate(p)$
        \STATE $updateBestScore(score,p)$
        \STATE $neighbours=getNeighbours(p,\delta)$
        \FOR{$p2:neighbours$}
            \STATE $enqueue(q,p2,score)$
        \ENDFOR
    
    \ENDWHILE
\ENDFOR

\RETURN $point^{*}$
\end{algorithmic}
\end{algorithm}

The experiments described in the next two sections showed that PQMeLiF performed much better than MeLiF and MeLiF+. However, they uncovered its drawbacks. The algorithm starts working initially with only the set of starting points in the queue, number of which is fixed and equals to the number of the basic ranking filters plus one. So, if a server has more cores, the extra cores stay unused until new points are added to the queue. The possible solution for this problem is to process more starting points in order to keep all the server cores busy. This algorithms drawback can be overcome with our next algorithm.

The main idea of this algorithm is to consider the problem of selecting new points as a reinforcement learning problem, in which we need to find a trade-off between exploration (of new areas in the search space) and exploitation (by evaluating points in areas where we have already found many good points)~\citep{sutton1998reinforcement}. In order to apply this idea, we adopted the well-known UCB1 algorithm~\citep{auer2002finite} for our parallelization problem by reducing it to multi-armed bandit problem. This reduction was performed in the following way. Firstly, we split the search space into different areas and correspond each area with an arm. Then, evaluating a point in an area is understood as playing the corresponding arm. Reward obtained by such playing is the score of the evaluated point~\citep{bubeck2012best,desautels2014parallelizing}.

\begin{algorithm}
\caption{MAMeLiF pseudo code}
\begin{algorithmic} 
\REQUIRE starting points, $\delta$, $evaluate$, $T$
\STATE $points=splitSearchSpace(P)$
\STATE $queues=[]$
\STATE $q^{*}$
\STATE $p^{*}$
\FOR {$p:points$}
    \STATE $q=PriorityBlockingQueue$
    \STATE $enqueue(q,p,1.0)$
    \STATE $queues+=q$
\ENDFOR

\FOR{$each\ new\ thread\ run\ in\ T\ threads$}
    \WHILE{$!mustStop()$}
        \STATE $qNE=findNonEmpty(queues)$
        \STATE $q=findBestQ(qNE)$
        \STATE $p=dequeue(q)$
        \STATE $score=evaluate(p)$
        \STATE $updateBestScore(score,p)$
        \STATE $neighbours=getNeighbours(p,\delta)$
        \FOR{$p2:neighbours$}
            \STATE $enqueue(q,p2,score)$
        \ENDFOR
    \ENDWHILE
\ENDFOR

\RETURN $point^{*}$
\end{algorithmic}
\end{algorithm}

The problem we faced during implementation of reinforcement learning algorithm in multi-agent environment is delayed feedback. Basically, when some thread needs to select which arm to use, there is no information about the results of other threads at that moment. So it is forced to make a decision with the lack of information. Authors of~\citep{joulani2013online} proposed to make a decision based only on results that were already computed and provide a theoretical proof, that an error function is additive and based on the duration of delay. 

\section{Experimental setup}

As a classifier, we used SVM with polynomial kernel and soft margin parameter $C = 1$ implemented in WEKA library. We used 5-fold cross-validation. The number of selected features was constant: $m=100$.

We ran our experiments on a machine with the following characteristics: 32-core CPU AMD Opteron 6272 @ 2.1 GHz, 128 GB RAM. We used $K=50$ threads, where  $K=2pf$, $p$ is the number of starting points, and $f$ is the number of folds.

As the basic filters, we used Spearman Rank Correlation, Symmetric Uncertainty, Fit Criterion, VDM~\citep{auffarth2010comparison}. We also executed MeLiF and MeLiF+ and recorded work time and point with the best classification result.

We used 36 datasets of different sizes from these archives: GEO, Broad institute. Cancer Program Data Sets, Kent Ridge Bio-Medical Dataset, Feature Selection Datasets at Arizona State University, RSCTC’2010 Discovery Challenge  All these datasets are DNA-microarray datasets with high number of features (from a few thousands up to a few dozens of thousands) and comparatively low number of objects (less than few hundreds). 

\section{Results}

We used several different configurations of halt criteria in our algorithms that described below. For each dataset, we conducted 10 experiments total:

\begin{itemize}
\item B: basic one-thread MeLiF;
\item P: naive parallelization method MeLiF+;
\item PQ75: PQMeLiF limited to visit 75 points;
\item PQ100: PQMeLiF limited to visit 100 points;
\item PQ125: PQMeLiF limited to visit 125 points;
\item PQrel: PQMeLiF that stops when no quality increase was registered in the previous 32 points;
\item MA75: MAMeLiF limited to visit 75 points;
\item MA100: MAMeLiF limited to visit 100 points;
\item MA125: MAMeLiF limited to visit 125 points;
\item MArel: MaMeLiF that stops when no quality increase was registered in the previous 32 points;
\end{itemize}

Each one of them halted if it gets the highest score $1.0$. The results of the basic MeLiF, MeLiF+ are presented in the Table~\ref{tab:Results}. Only best configurations for PQMeLiF and MAMeLiF are presented in it, that are PQ75 and MArel correspondingly. As it could be seen from the Table~\ref{tab:Results}, PQMeLiF and MAMeLiF strongly outperform MeLiF+ resulting in approximately linear scaling over the number of computational cores. Also new methods gain in average small boost to the original MeLiF feature selection quality. 

\begin{longtable}{c|cccc|cccc}

  \caption{Algorithms comparison}
  \label{tab:Results} \\

\hline

{\multirow{2}{*}{Dataset}} & \multicolumn{4}{c|}{Time} & \multicolumn{4}{c}{$F_{1}$ score} \\ \cline{2-9} 
& MeLiF \hphantom{s} & MeLiF+\hphantom{s}  & PQ75 & MArel & 
{MeLiF} & MeLiF+ & PQ75 & MArel \\

    \hline
    
Arizona1 & 558 & 85 & 85 & 117 & 0.833 & 0.833 & 0.833 & 0.833 \\
Arizona5 & 219 & 67 & 37 & 48 & 0.768 & 0.79 & 0.786 & 0.773 \\
Breast & 161 & 24 & 27 & 17 & 0.844 & 0.822 & 0.812 & 0.802 \\
CNS & 33 & 6 & 7 & 6 & 0.742 & 0.791 & 0.899 & 0.83 \\
Data\_train0 & 172 & 71 & 28 & 32 & 0.853 & 0.853 & 0.849 & 0.839 \\
Data\_train1 & 180 & 60 & 32 & 33 & 0.866 & 0.901 & 0.877 & 0.876 \\
Data4\_train & 513 & 124 & 73 & 87 & 0.823 & 0.823 & 0.775 & 0.775 \\
Data5\_train & 370 & 70 & 59 & 58 & 0.847 & 0.847 & 0.901 & 0.886 \\
Data6\_train & 381 & 69 & 65 & 64 & 0.835 & 0.835 & 0.859 & 0.869 \\
DLBCL & 65 & 13 & 12 & 19 & 0.799 & 0.734 & 0.8 & 0.761 \\
GDS2771 & 299 & 81 & 42 & 42 & 0.798 & 0.798 & 0.801 & 0.783 \\
GDS2819\_1 & 303 & 39 & 15 & 17 & 1 & 1 & 1 & 1 \\
GDS2819\_2 & 436 & 149 & 60 & 80 & 0.948 & 0.981 & 0.957 & 0.921 \\
GDS2901 & 88 & 17 & 4 & 4 & 1 & 1 & 1 & 1 \\
GDS2960 & 33 & 7 & 5 & 6 & 0.99 & 0.99 & 0.977 & 0.977 \\
GDS2961 & 49 & 13 & 8 & 5 & 0.86 & 0.86 & 0.829 & 0.784 \\
GDS2962 & 45 & 11 & 8 & 7 & 0.877 & 0.914 & 0.924 & 0.883 \\
GDS3116 & 142 & 23 & 30 & 32 & 0.852 & 0.852 & 0.868 & 0.853 \\
GDS3257 & 131 & 17 & 8 & 9 & 1 & 1 & 1 & 1 \\
GDS3929	& 376 & 74 & 45 & 32 & 0.809 & 0.809 & 0.81 & 0.774 \\
GDS4103 & 265 & 71 & 54 & 51 & 0.933 & 0.933 & 0.923 & 0.923 \\
GDS4109 & 142 & 38 & 24 & 20 & 0.936 & 0.936 & 0.924 & 0.947 \\
GDS4222 & 454 & 84 & 73 & 93 & 0.974 & 0.974 & 0.97 & 0.97 \\
GDS4318 & 275 & 64 & 40 & 53 & 0.923 & 0.923 & 0.97 & 0.942 \\
GDS4336 & 200 & 66 & 30 & 20 & 0.928 & 0.928 & 0.916 & 0.916 \\
GDS4431 & 537 & 100 & 85 & 134 & 0.827 & 0.827 & 0.817 & 0.817 \\
GDS4600 & 472 & 124 & 94 & 114 & 0.983 & 0.983 & 0.979 & 0.979 \\
GDS4837\_1 & 413 & 130 & 57 & 51 & 0.916 & 0.916 & 0.828 & 0.809 \\
GDS4837\_3 & 316 & 48 & 54 & 51 & 0.96 & 0.96 & 0.969 & 0.967 \\
GDS4901 & 220 & 60 & 44 & 30 & 0.931 & 0.966 & 0.919 & 0.913 \\
GDS4968\_0 & 226 & 40 & 37 & 38 & 0.905 & 0.905 & 0.913 & 0.907 \\
GDS4968\_1 & 224 & 40 & 40 & 36 & 0.923 & 0.946 & 0.939 & 0.932 \\
GDS5037\_0 & 243 & 140 & 49 & 62 & 0.825 & 0.857 & 0.867 & 0.867 \\
GDS5037\_2 & 293 & 69 & 47 & 73 & 0.756 & 0.756 & 0.789 & 0.78 \\
GDS5047 & 185 & 41 & 9 & 11 & 1 & 1 & 1 & 1 \\
GDS5083 & 195 & 60 & 29 & 40 & 0.862 & 0.862 & 0.872 & 0.847 \\
Leuk\_3c0	& 34 & 5 & 7 & 7 & 0.989 & 0.989 & 0.986 & 0.986 \\
Leuk\_3c1 & 33 & 5 & 8 & 8 & 0.981 & 0.981 & 0.98 & 0.98 \\
Ovarian & 192 & 23 & 9 & 11 & 1 & 1 & 1 & 1 \\
plySRBCT & 17 & 3 & 0 & 1 & 1 & 1 & 1 & 1 \\
prostate & 93 & 34 & 16 & 15 & 0.919 & 0.932 & 0.927 & 0.921 \\

    \hline
%
\end{longtable}

\section{Conclusion}
In this paper, we presented two parallelization schemes for MeLiF algorithm, which search for the best combination of simple feature selection algorithms. Experiments showed that these two schemes, namely PQMeLiF and MAMeLiF, demonstrated linear speed improvement over the number of used cores comparing to single-threaded MeLiF without loss in classification quality. Furthermore, these algorithms sometimes showed better feature selection quality. This could be explained with the fact, that this methods search through more points than original MeLiF method. That happens due to delay in thread synchronization. 

As our future research, we will try to make some estimations on search space split size in MAMeLiF method depending on the number of cores. Proper split can lead to better computational results. Also, we will apply more tests with different system configurations in order to find Amdahl’s optimum for each algorithm.

\acks{ The research was supported by the Government of the Russian Federation (grant~074-U01) and the Russian Foundation for Basic Research (project no. 16-37-60115). }

\bibliography{acml16}

\end{document}